\documentclass[letterpaper, 10 pt, conference]{ieeeconf}  

\IEEEoverridecommandlockouts                              

\usepackage{graphics} 
\usepackage{epsfig} 
\usepackage{mathptmx} 
\usepackage{times} 
\usepackage{amsmath} 
\usepackage{amssymb}  

\usepackage[utf8]{inputenc}
\usepackage[noend]{algpseudocode}
\usepackage[utf8]{inputenc}
\usepackage{xcolor}
\usepackage{graphicx}
\usepackage{hyperref}
\usepackage{booktabs}
\usepackage{color}
\usepackage{array}
\usepackage{booktabs}
\usepackage{multirow}
\usepackage{geometry}
\usepackage{hyperref}
\usepackage[space]{cite}
\usepackage[linesnumbered,boxed,ruled,commentsnumbered]{algorithm2e}
\usepackage[utf8]{inputenc}
\usepackage[noend]{algpseudocode}
\usepackage{amsmath}
\usepackage{amssymb}    
\usepackage[utf8]{inputenc}
\usepackage{xcolor}
\usepackage{graphicx}
\usepackage{hyperref}
\usepackage{booktabs}
\usepackage{color}
\usepackage{array}
\usepackage{booktabs}
\usepackage{multirow}
\usepackage{geometry}
\usepackage{hyperref}
\usepackage[space]{cite}
\usepackage{mathtools}
\usepackage{threeparttable} 
\usepackage{makecell} 
\usepackage{wasysym}
\usepackage{lipsum}
\usepackage{tabularx} 
\usepackage{makecell} 
\DeclareGraphicsExtensions{.pdf,.png,.jpg,.eps}
\title{\LARGE \bf
FIRE-LIVWO: Robust LiDAR-Inertial-Visual-Wheel Odometry via Failure-Immune mmWave Radar Enhancement
}

\begin{document}

\author{Kun Hu$^{1}$,
    Menggang Li$^{1,2,3,\dagger}$,
    Kaidi Wu$^{1}$,
    Zhiwen Jin$^{1}$, \\
    Yingjie Zhao$^{1}$,
    Chaoquan Tang$^{1}$,
    Eryi Hu$^{1}$ and Gongbo Zhou$^{1}$
    \thanks{Research supported by the National Natural Science Foundation of China
    (grant number: 52304183, 52274159), the Jiangsu Provincial Key Research and Development (R\&D) Plan Projects (BE2023008-4), and the Project Funds of the Priority Academic Program Development of Jiangsu Higher Education Institutions (PAPD).}%
    \thanks{$^{1}$K. Hu, M. Li, K. Wu, Z. Jin, Y. Zhao, C. Tang, G. Zhou and E. Hu are with the School of Mechatronic Engineering, China University of Mining and Technology.
        {\tt\footnotesize {ts22050017a31, sallylmg, ts23050205p31, ts24050202p31, ts23050157p31, tangchaoquan, gbzhou}@cumt.edu.cn, horyhu@126.com}}%
    \thanks{$^{2}$M. Li is with the Jiangsu Collaborative Innovation Center of
    Intelligent Mining Equipment, China University of Mining and Technology.}%
    \thanks{$^{3}$M. Li is with the National Key Laboratory of Intelligent Mining Equipment Technology, China University of Mining and Technology.}%
    \thanks{Kun Hu and Menggang Li contributed equally to this work and should be considered co-first authors.}%
    \thanks{$^{\dagger}$Menggang Li is the corresponding author.}%
}



\maketitle
\thispagestyle{empty}
\pagestyle{empty}

\begin{abstract}
Achieving robust simultaneous localization and mapping (SLAM) in large-scale underground coal mines with complex structures and severe degeneracies remains highly challenging. Dense smoke and dust often cause substantial loss of visual information and degrade LiDAR point-cloud features, while long, self-similar corridors induce geometric degeneration that leads to pronounced odometry drift along specific degrees of freedom. To address these issues, we propose \textbf{FIRE-LIVWO: Failure-Immune mmWave Radar-Enhanced LiDAR–Inertial–Visual–Wheel Odometry}, a tightly coupled multi-modal odometry framework based on an iterated error-state Kalman filter (IESKF). The framework fuses 4D millimeter-wave (mmWave) radar, LiDAR, and visual features within a unified VoxelMap and jointly constructs LiDAR–radar point-to-plane residuals and sparse visual photometric residuals for the filter update. In smoke-filled environments, we exploit the strong penetration capability of 4D mmWave radar and introduce pointwise Doppler velocity constraints to preserve state observability. In geometrically degenerate corridors, we tightly couple wheel odometry using non-holonomic constraints (NHC) and online lever-arm compensation to reduce drift. Our central contribution is a degeneration detection and adaptive fusion model switching strategy grounded in geometric and visual observability analysis, which quantifies observability online and dynamically adjusts the weights and activation states of each modality. Real-world experiments in underground coal mines demonstrate that FIRE-LIVWO accurately identifies the boundaries of visual failures and geometric underconstraints, enabling reliable modality switching under extreme conditions. Compared with baseline methods, it achieves superior accuracy and robustness, with an average localization error of 5.677~m. We open source our code of this work on Github\footnote[3]{\url{https://github.com/KJ-Falloutlast/FIRE-LIVWO}} to benefit the robotics community.
\end{abstract}

\section{INTRODUCTION}

Underground coal-mine tunnels feature long travel distances, poor illumination, heavy dust and smoke, and highly repetitive layouts. In post-disaster rescue operations, visibility can deteriorate abruptly, and the ground may become uneven and unstructured, creating substantial hazards for mobile robots performing autonomous inspection and search-and-rescue. Building high-precision, long-term digital-twin colored point-cloud maps in such environments typically relies on LiDAR--Visual--Inertial (LVI) SLAM systems. However, in complex underground mines, LVI systems often fail due to the combined effects of heterogeneous degeneration mechanisms: (1) \emph{Visual degradation:} high-concentration smoke reduces image contrast and disrupts photometric consistency; (2) \emph{Geometric degradation:} long, straight corridors and smooth walls provide insufficient geometric structure in LiDAR point clouds, causing pronounced drift in state estimation; and (3) \emph{Limitations of multi-modal fusion:} existing degeneration-handling methods often rely on a single Hessian-threshold test or fixed sensor-fusion strategies, limiting their ability to adaptively schedule multi-source information under mixed degeneracies.

To address these challenges, we present FIRE-LIVWO, a robust, tightly coupled multi-modal odometry system based on IESKF. Our main contributions are as follows:

\begin{itemize}
    \item \textbf{Contribution 1:} We propose a tightly-coupled framework that jointly optimizes LiDAR--radar geometric residuals and visual photometric residuals within a unified VoxelMap. The framework further integrates radar Doppler velocity constraints to improve robustness and prevent divergence in smoke-filled environments.
    \item \textbf{Contribution 2:} To mitigate geometric degradation in long underground corridors, we introduce a tightly coupled wheel-odometry module that incorporates non-holonomic constraints (NHC) and adaptive online lever-arm compensation, thereby reducing corridor-induced drift.
    \item \textbf{Contribution 3:} We design a degradation detection and adaptive fusion strategy based on geometric and visual observability analysis. By monitoring observability scores online, the system identifies smoke interference and geometric underconstraints, dynamically switches fusion modalities, and maintains continuous state estimation across scenarios.
    \item \textbf{Contribution 4:} Extensive real-world experiments in underground coal mines validate the proposed method. We also conduct comprehensive evaluations against state-of-the-art approaches, demonstrating superior accuracy and reliability under challenging conditions.
\end{itemize}


\section{Related Work}\label{chap2-Related_work}
Simultaneous localization and mapping (SLAM) in large-scale, unknown, and structurally complex underground coal mines remains highly challenging. Low illumination, pervasive dust and smoke, and long, self-similar corridor geometries make it difficult for conventional SLAM systems that rely on a single exteroceptive modality to operate reliably over long durations in real deployments. Prior work has primarily improved robustness along two directions: (i) tightly coupled multi-sensor fusion to increase measurement redundancy and enhance state observability; and (ii) degeneration detection with adaptive constraint scheduling to avoid incorporating unreliable measurements when observations are degraded.

Tightly coupled LiDAR--inertial odometry (LIO), which combines LiDAR geometric constraints with inertial measurements from an IMU, is a mainstream approach for underground localization. LIO-SAM~\cite{shan2020lio} and FAST-LIO2~\cite{xu2022fast} achieve tight coupling of LiDAR and IMU measurements. To mitigate LiDAR performance degradation in weakly structured scenes (e.g., long, straight corridors), researchers have developed LiDAR--visual--inertial (LVI) systems that incorporate vision. LVI-SAM~\cite{shan2021lvi} integrates LIO and VIO in a factor-graph framework to improve robustness, while R3LIVE~\cite{lin2022r}, FAST-LIVO~\cite{zheng2022fast}, FAST-LIVO2~\cite{zheng2024fast}, and SR-LIVO~\cite{yuan2024sr} adopt an iterated error-state Kalman filter (IESKF) to tightly couple LiDAR, visual, and inertial data.

In recent years, 4D mmWave radar has attracted increasing attention because of its strong penetration capability in dust and smoke. 4DRadarSLAM~\cite{zhang20234dradarslam} presented an end-to-end mmWave SLAM system. GaRLIO~\cite{noh2025garlio} proposed a gravity-aided radar--LiDAR--inertial odometry method that leverages pointwise Doppler velocities to improve gravity estimation and reduce vertical drift. DR-LRIO~\cite{nissov2024degradation} introduced a degradation-robust LiDAR--radar--inertial odometry system and demonstrated robust localization in degenerate environments, including geometrically self-similar tunnels and dense fog.

Wheel odometry provides proprioceptive motion information that is independent of exteroceptive sensing and is therefore valuable in geometrically degenerate settings such as tunnels. Mandow \emph{et al.}~\cite{mandow2007experimental} proposed an extended differential-drive model. ACK-MSCKF~\cite{ma2019ack} preintegrates wheel-odometry forward-velocity and angular-rate measurements, but assumes flat terrain. Liu \emph{et al.}~\cite{liu2019visual} improved localization accuracy by preintegrating encoder and gyroscope measurements.

Degradation detection is essential for robust operation in complex environments. Zhang \emph{et al.}~\cite{zhang2016degeneracy} detected odometry degradation by analyzing the eigenvalues of the LiDAR Hessian matrix. X-ICP~\cite{tuna2023x} combined localizability detection with optimization to improve scan-matching accuracy under degradation. LIVER~\cite{wen2024liver} integrated LiDAR degradation handling with learning-based image enhancement. However, these systems rely on MAP-based fusion; when sensors experience long-duration failures, the estimator may diverge. Moreover, they can be overly dependent on vision and therefore may not operate reliably in no-illumination scenarios.

\section{Preliminary}\label{chap3-Preliminary}
\subsection{Notations and Definitions}


\begin{table}[h]
    \centering
    \caption{Definitions of Important Variables}
    \label{tbl1:symbols}
    \begin{tabularx}{\columnwidth}{lX}
        \toprule
        Notation & Meaning \\
        \midrule
        $\mathbf{x},\hat{\mathbf{x}},\overline{\mathbf{x}}$ 
            & Ground truth, predicted, and updated estimates of $\mathbf{x}$. \\
        $\delta\mathbf{x}$ 
            & Error-state vector. \\
        $^{G}(\cdot),\, ^{B}(\cdot)$ 
            & Vector in global or body frame. \\
        $^{R}(\cdot),\, ^{L}(\cdot),\, ^{C}(\cdot)$ 
            & Vector in radar, LiDAR, or camera frame. \\
        $^{W}(\cdot),\, ^{I}(\cdot)$ 
            & Vector in wheel or IMU frame. \\
        \bottomrule
    \end{tabularx}
\end{table}

We assume that the temporal offsets among the five sensors—LiDAR, IMU, camera, wheel odometer, and 4D mmWave radar—are known (via pre-calibration or synchronization). The IMU frame, denoted by $I$, is used as the body frame, and the global frame, denoted by $G$, is defined at the origin of the total-station coordinate system (Table~\ref{tbl1:symbols}). All sensors are rigidly mounted, and the LiDAR, camera, and IMU are hardware-synchronized. The discrete-time state transition model at the $i$-th IMU measurement is
\begin{equation}
    \mathbf{x}_{i+1}=\mathbf{x}_i \boxplus\left(\Delta t\, \mathbf{f}\left(\mathbf{x}_i, \mathbf{u}_i, \mathbf{w}_i\right)\right),
    \label{eq_1_state_propagate}
\end{equation}
where $\boxplus/\boxminus$ denote the generalized addition/subtraction operators~\cite{xu2022fast}, $\Delta t$ is the IMU sampling period, and the state $\mathbf{x}\in\mathbb{R}^{18}$ is defined as
\begin{equation}
    \mathbf{x} \triangleq\left[\begin{array}{llllll}
    { }^G \mathbf{R}_I^T & { }^G \mathbf{p}_I^T & { }^G \mathbf{v}^T & \mathbf{b}_{\mathbf{g}}^T & \mathbf{b}_{\mathbf{a}}^T & { }^G \mathbf{g}^T 
    \end{array}\right]^T.
    \label{eq_2_state}
\end{equation}
The input $\mathbf{u}$, process noise $\mathbf{w}$, and function $\mathbf{f}$ follow the definitions in~\cite{xu2022fast}; due to space limitations, we omit the details here.

\subsection{Error-State Iterated Kalman Filter Update}

The propagated state $\hat{\mathbf{x}}_k$ and covariance $\hat{\mathbf{P}}_k$ obtained from forward propagation~\cite{xu2022fast} impose a prior distribution on $\mathbf{x}_k$:
\begin{equation}
    \mathbf{x}_k \boxminus \hat{\mathbf{x}}_k \sim \mathcal{N}\left(\mathbf{0}, \hat{\mathbf{P}}_k\right).
    \label{eq_3_prior_distribution}
\end{equation}
Combining the prior in (\ref{eq_3_prior_distribution}) with radar Doppler-velocity measurements $\mathbf{z}_{R_V}$, radar geometric measurements $\mathbf{z}_{R_G}$, LiDAR measurements $\mathbf{z}_{L}$, visual measurements $\mathbf{z}_{C}$, and wheel-odometry measurements $\mathbf{z}_{W}$, we obtain the maximum a posteriori (MAP) estimate of $\delta \mathbf{x}_k$:
\begin{equation}
    \begin{aligned}
    \min _{\delta \mathbf{x}_k \in \mathcal{M}} \Bigg( 
        & \left\|\mathbf{x}_k \boxminus \hat{\mathbf{x}}_k\right\|_{\hat{\mathbf{P}}_k}^2 
        + \sum_{i=1}^{N_{R_V}} \left\|\mathbf{r}_{R_V}\!\left(\mathbf{z}_{R_V}^i, \mathbf{x}_k\right)\right\|_{\mathbf{P}_{R_V}^i}^2 \\
        & + \sum_{i=1}^{N_{R_G}} \left\|\mathbf{r}_{R_G}\!\left(\mathbf{z}_{R_G}^i, \mathbf{x}_k\right)\right\|_{\mathbf{P}_{R_G}^i}^2 
        + \sum_{i=1}^{N_{L}} \left\|\mathbf{r}_{L}\!\left(\mathbf{z}_{L}^i, \mathbf{x}_k\right)\right\|_{\mathbf{P}_{L}^i}^2 \\
        & + \sum_{i=1}^{N_{C}} \left\|\mathbf{r}_{C}\!\left(\mathbf{z}_{C}^i, \mathbf{x}_k\right)\right\|_{\mathbf{P}_{C}^i}^2 
        + \sum_{i=1}^{N_{W}} \left\|\mathbf{r}_{W}\!\left(\mathbf{z}_{W}^i, \mathbf{x}_k\right)\right\|_{\mathbf{P}_{W}^i}^2 \Bigg),
    \end{aligned}
    \label{eq_4_constrains}
\end{equation}
where $\|\mathbf{x}\|_{\mathbf{P}}^2 \triangleq \mathbf{x}^T \mathbf{P}^{-1} \mathbf{x}$, and $\mathbf{r}_{R_V}$, $\mathbf{r}_{R_G}$, $\mathbf{r}_{L}$, $\mathbf{r}_{C}$, and $\mathbf{r}_{W}$ denote the Doppler-velocity residual, radar geometric residual, LiDAR residual, visual residual, and wheel-odometry velocity residual, respectively. $\mathbf{P}_{R_V}^i$, $\mathbf{P}_{R_G}^i$, $\mathbf{P}_{L}^i$, $\mathbf{P}_{C}^i$, and $\mathbf{P}_{W}^i$ are the corresponding measurement covariance matrices, and $N_{R_V}$, $N_{R_G}$, $N_{L}$, $N_{C}$, and $N_W$ are the numbers of measurements within the interval from $t_{k-1}$ to $t_k$. The optimization problem in (\ref{eq_4_constrains}) is non-convex and can be solved iteratively using the Gauss--Newton method; this iterative optimization is equivalent to an iterated Kalman filter~\cite{bell1993iterated}.

\section{System Overview}\label{chap4-system_overview}
\begin{figure}
    \centering
    \includegraphics[width=1.0\columnwidth]{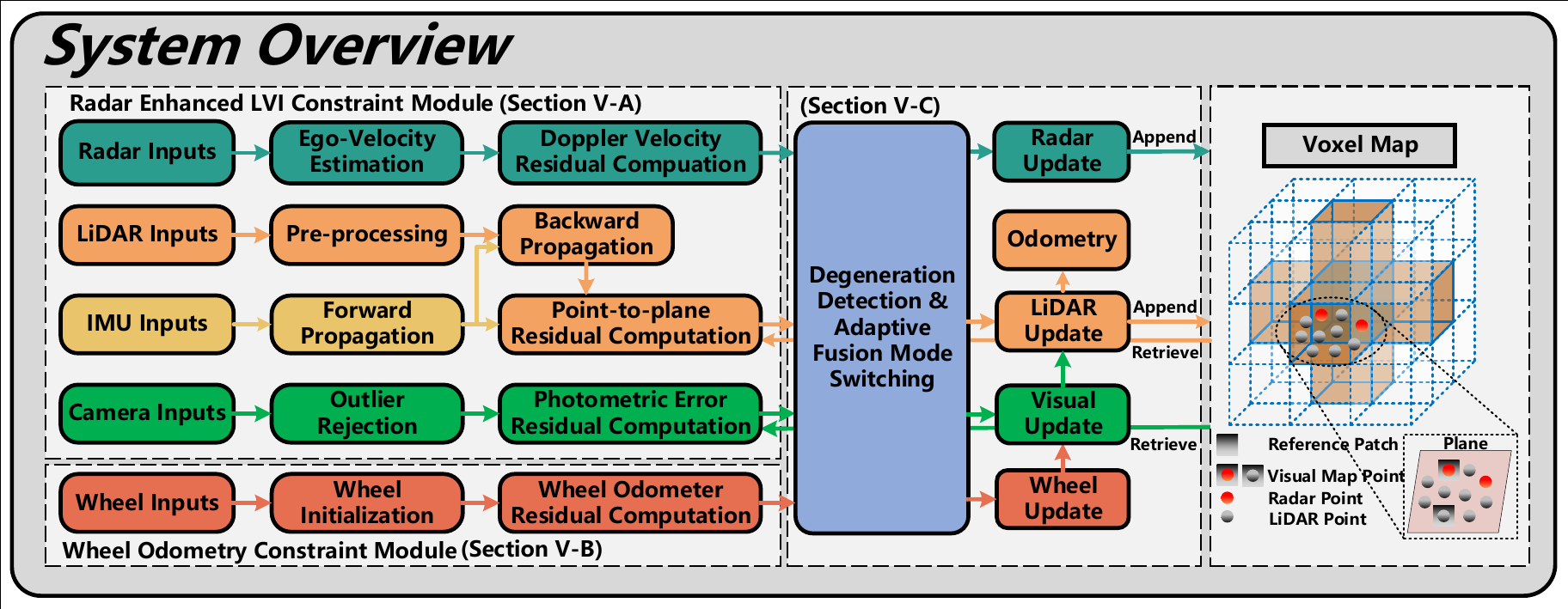}
    \caption{System overview.}
    \label{Fig1-System overview}
\end{figure}

Our objective is to estimate the 6-DoF pose of a coal-mine robot while constructing a global map. The overall system architecture is shown in Fig.~\ref{Fig1-System overview}. The system comprises three modules:

\textbf{(i) Radar-Enhanced LiDAR Visual Inertial Constraint Module:} We extract point clouds from mmWave radar echoes. Using ego-velocity estimation, we reject outliers and construct Doppler-velocity constraints (Chapter~\ref{chap5-1-radar-enhanced-module}). Radar points are then indexed into the map and inserted into a unified VoxelMap jointly maintained with LiDAR and visual data, where radar and LiDAR point-to-plane residuals are formulated in a consistent manner (Chapter~\ref{chap5-1-2-unified-residuals}). In addition, radar and LiDAR points are projected onto the image plane to construct sparse direct photometric errors (Chapter~\ref{chap5-1-3-visual_constrains}). By adopting a unified map representation, this module improves robustness under severe smoke and dust conditions.

\textbf{(ii) Vehicle Kinematic Constraint Module:} We obtain velocity measurements from wheel odometry, synchronize wheel-speed data with LiDAR timestamps, and formulate vehicle kinematic constraints by incorporating non-holonomic constraints (NHC) and online lever-arm compensation (Chapter~\ref{chap5-2-wheel_constrains}).

\textbf{(iii) Degradation Detection and Adaptive Fusion Mode Switching Module:} This module assesses the reliability of multi-source observations in real time and adaptively schedules the fusion strategy. We design a visual degradation detector based on an atmospheric scattering model and a geometric degradation detector based on eigenvalue analysis of the geometric Hessian matrix. The detectors compute a visual observability score $\mathcal{O}_V$ and a geometric observability score $\mathcal{O}_G$, respectively. These indicators identify heterogeneous degeneration boundaries and dynamically adjust the activation of residual terms in the IESKF, enabling robust state estimation in complex degenerate scenarios (Chapter~\ref{chap5-3-degradation_detection}).

\section{Methodology}\label{chap5-Methodology}
\subsection{Radar-Enhanced LiDAR-Visual-Inertial Constraint Module}\label{chap5-1-radar-enhanced-module}

\subsubsection{Unified VoxelMap and Visual Map-Point Selection Strategy}\label{chap5-1-1-unified-voxelmap}

We employ an adaptive, octree-based voxel map (Fig.~\ref{Fig1-System overview}) with a root voxel size of 0.5~m. Each leaf voxel stores planar features $\Pi \triangleq (\mathbf{n}, \mathbf{q}, \boldsymbol{\Sigma}_{\mathbf{n},\mathbf{q}})$ and LiDAR points $^{L}\mathbf{p}_i$. A subset of points is augmented with $8\times 8$ image patches to form a set of visual map points, denoted by $\mathcal{P}^{map}$: converged points retain only the reference patch, whereas unconverged points additionally store other visible patches. Visual map points are drawn from two sources: LiDAR candidate points lying on planes, denoted by $\mathcal{P}^{map}_L$, and radar candidate points lying on planes, denoted by $\mathcal{P}^{map}_R$. Accordingly, $\mathcal{P}^{map}=\mathcal{P}^{map}_L\cup\mathcal{P}^{map}_R$. To ensure robust visual alignment, we adopt the map-point selection strategy in~\cite{zheng2024fast} and extend it to (i) explicitly incorporate radar field-of-view constraints and (ii) support heterogeneous point sources. This design helps maintain stable visual alignment when LiDAR measurements become degenerate.

\subsubsection{Unified Geometric Residuals for LiDAR/Radar and Doppler-Velocity Residuals}\label{chap5-1-2-unified-residuals}
We propose a dual-constraint strategy that fuses LiDAR and 4D mmWave radar feature points by jointly leveraging (i) geometric constraints from both LiDAR and radar points and (ii) Doppler-velocity constraints from radar points. In dense smoke, where LiDAR point-cloud features can degrade, radar geometric constraints help preserve reliable data association, while Doppler-velocity residuals provide complementary motion information. Together, these constraints improve estimation accuracy and robustness.

\textbf{Geometric residuals of LiDAR and radar points:}
Following~\cite{xu2022fast}, upon receiving a LiDAR scan at time $t_k$, we first compensate motion distortion via backward propagation so that each point ${}^L\mathbf{p}_j$ can be treated as being sampled at $t_k$. We then construct a point-to-plane observation model. If the LiDAR point ${}^L\mathbf{p}_j$ is transformed into the global frame using the true state $\mathbf{x}_k$, its point-to-plane residual should be zero:
\begin{equation}
    \mathbf{0}=\mathbf{r}_L^j\left(\mathbf{x}_k, \mathbf{n}_j^L\right)
    =\mathbf{u}_j^T\left({ }^G \mathbf{T}_{I_k}{ }^I \mathbf{T}_L{ }^L \mathbf{p}_j-\mathbf{q}_j\right),
    \label{eq_5_lidar_constrains}
\end{equation}
where $\mathbf{n}_j^L$ denotes the LiDAR measurement noise, and $\mathbf{u}_j$ and $\mathbf{q}_j$ are the normal vector and a point on the matched plane in the map, respectively. Similarly, for a radar feature point ${}^R\mathbf{p}_j$, the radar geometric residual $\mathbf{r}_{R_G}^j$ is defined as
\begin{equation}
    \mathbf{0}=\mathbf{r}_{R_G}^j\left(\mathbf{x}_k, \mathbf{n}_j^{R_G}\right)
    =\mathbf{u}_j^T\left({ }^G \mathbf{T}_{I_k}{ }^I \mathbf{T}_R {}^R\mathbf{p}_j-\mathbf{q}_j\right),
    \label{eq_6_radar_constrains}
\end{equation}
where $\mathbf{n}_j^{R_G}$ denotes the radar geometric measurement noise, and ${ }^I \mathbf{T}_R$ is the radar-to-IMU extrinsic parameter. Equations~(\ref{eq_5_lidar_constrains}) and~(\ref{eq_6_radar_constrains}) define the LiDAR residual and the radar geometric residual, respectively.

\textbf{Doppler-velocity residuals of radar points:}
We further use radar points ${}^{R}\mathbf{p}_j$ to construct Doppler-velocity residuals. The linear velocity expressed in the radar frame is defined as
\begin{equation}
    \mathbf v_R(\mathbf x_k, \mathbf{n}_j^{R_V}) \triangleq \mathbf v_R
    = {}^I\mathbf R_R^\top \Big( {}^G\mathbf R_{I_k}^\top\,{}^G\mathbf v_{I_k}
    +\lfloor \hat\omega_{I}\rfloor_\times\,{}^I\mathbf p_R \Big),
    \label{eq_7_radar_ego_velocity}
\end{equation}
which leads to the Doppler-velocity residual constraint $\mathbf{r}_{R_V}$:
\begin{equation}
    \begin{aligned}
    \mathbf{r}_{R_V}(\mathbf{x}_k, \mathbf{n}_j^{R_V})
    &= \mathbf u({}^R\mathbf p_j)^\top\, {}^I\mathbf R_R^\top
    \Big( {}^G\mathbf R_{I_k}^\top\,{}^G\mathbf v_{I_k}
    +\lfloor \hat\omega_{I}\rfloor_\times\,{}^I\mathbf p_R \Big) - \hat v_{R_j},
    \end{aligned}
    \label{eq_8_radar_doppler_constrains}
\end{equation}
where ${ }^I\mathbf{R}_{R}$ and ${ }^I\mathbf{p}_{R}$ are the radar--IMU extrinsics, $\hat v_{R_j}$ is the Doppler measurement of the 4D mmWave radar at ${}^{R}\mathbf{p}_j$, $\lfloor \hat\omega_{I}\rfloor_\times$ is the skew-symmetric matrix of the angular velocity, and $\mathbf{u}(\cdot)$ denotes the unit direction vector $\mathbf u({}^R\mathbf p_j)=\frac{{}^R\mathbf p_j}{\|{}^R\mathbf p_j\|}$.

\subsubsection{Sparse Direct Photometric Residuals}\label{chap5-1-3-visual_constrains}

Inspired by~\cite{zheng2024fast}, we construct the visual measurement model using visual map points ${}^{G}\mathbf{p}_i \in \mathcal{P}^{map}$, where ${}^{G}\mathbf{p}_i$ may originate from either LiDAR or radar. The underlying assumption is that, after transforming the map point ${}^{G}\mathbf{p}_i$ into the current image $I_k(\cdot)$ using the true state $\mathbf{x}_k$, the photometric error between the reference patch and the current patch should be zero:
\begin{equation}
    \mathbf{r}_C(\mathbf x_k,\mathbf n_c)= \left( I_k(\mathbf{u}_i + \Delta \mathbf{u}) - \delta \mathbf{n}_{I_k} \right) -  \left( I_r(\mathbf{u}_i' + \mathbf{A}_i^r \Delta \mathbf{u}) - \delta \mathbf{n}_{I_r} \right),
    \label{eq_11_visual_residual}
\end{equation}
where $\mathbf{u}_i = \pi \left( {}^{{C}}\mathbf{T}_{{I}} \left( {}^{{G}}\hat{\mathbf{T}}_{{I}_\kappa} \right)^{-1} {}^{{G}}\mathbf{p}_i \right)$ and $\mathbf{u}_i' = \pi \left( {}^{{C}_r}\mathbf{T}_{{G}} \mathrm{Exp}(\delta \mathbf{T})\, {}^{{G}}\mathbf{p}_i \right)$ are the projections of the map point into the current and reference images, respectively. Here, $\pi(\cdot)$ denotes the camera projection model, $\Delta \mathbf{u}$ is the pixel offset within the patch, $\mathbf{A}_i^r$ is the affine transformation matrix, and $\mathbf{n}_c$ denotes image noise.

\subsection{Vehicle Kinematic-Constraint Module}\label{chap5-2-wheel_constrains}

In geometrically degenerate environments, such as long, self-similar tunnels, LiDAR or visual constraints can become under-constrained along certain directions, leading to accumulated drift. Wheel odometry provides proprioceptive motion information that is independent of external features and can therefore serve as an effective constraint under such structural degradation. Inspired by~\cite{hu2025cm}, we formulate the wheel-odometry constraint as
\begin{equation}
    \mathbf{r}_W(\mathbf x_k, \mathbf n_{W})={}^W\hat{\mathbf{v}}-{}^I\mathbf{R}_W^T \big({}^G\mathbf{R}_I^T\, {}^G\mathbf{v} + {}^I\boldsymbol{\omega} \times {}^I\mathbf{p}_W\big),
    \label{eq_12_wheel_residual}
\end{equation}
where ${}^W\hat{\mathbf{v}}$ is the velocity measurement expressed in the wheel-odometer frame, ${}^I\mathbf{v}$ and ${}^W\mathbf{v}$ denote the velocity expressed in the IMU and wheel frames, respectively, ${}^I\mathbf{R}_W$ and ${}^I\mathbf{p}_W$ are the wheel-to-IMU extrinsic parameters, ${}^I\boldsymbol{\omega}$ is the IMU angular velocity, and $\mathbf{n}_W$ denotes the measurement noise.

\subsection{Degradation Detection and Adaptive Fusion Mode Switching Module}\label{chap5-3-degradation_detection}
\begin{figure}[t]
\centering
    \includegraphics[width=1\columnwidth]{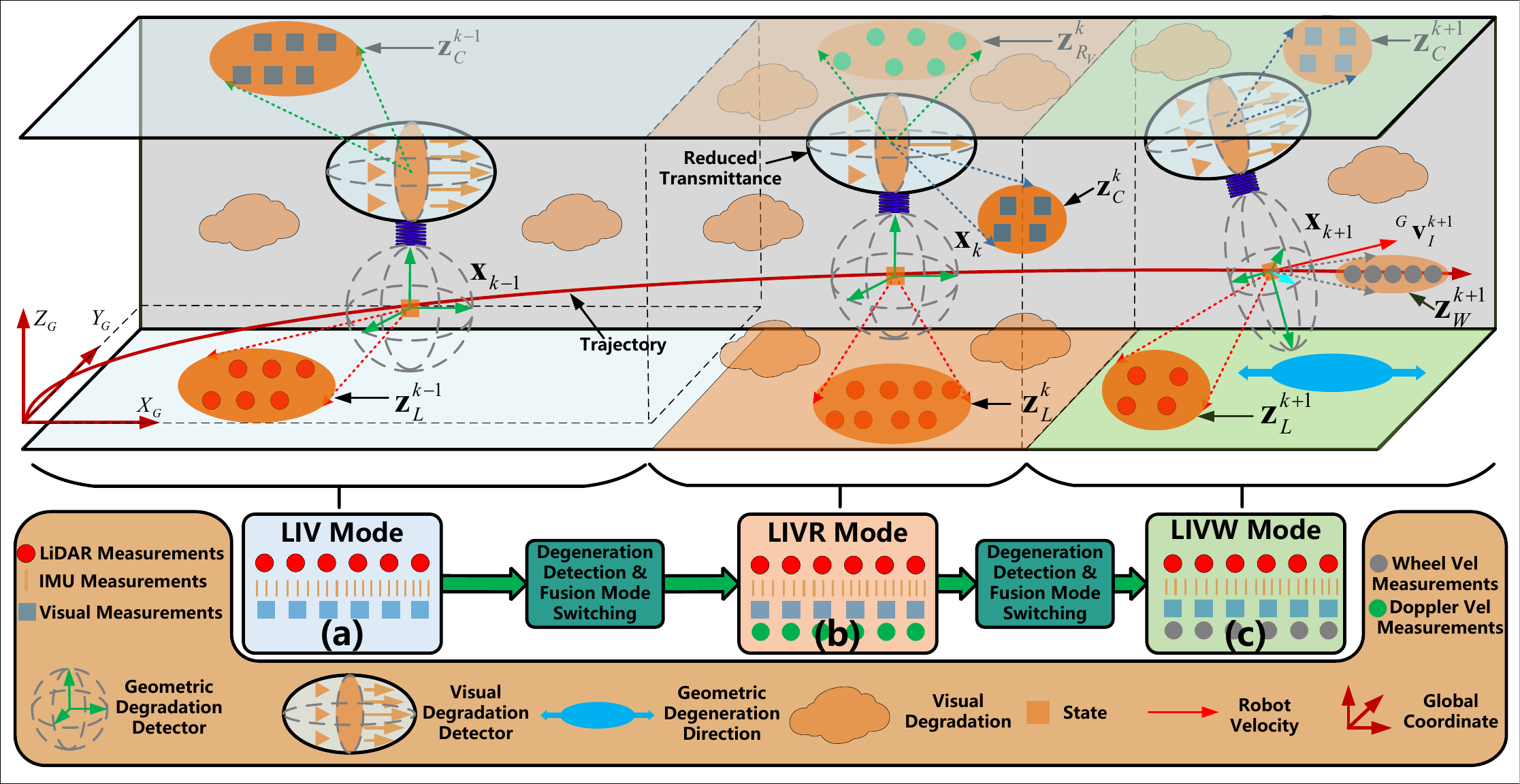}
    \caption{Degradation detection and adaptive fusion based on geometric and visual observability analysis. (a) In feature-rich corridor scenes, the system localizes using the LiDAR+IMU+Visual (LIV) mode, and the ellipsoid defined by the eigenvectors of the geometric Hessian matrix $\mathbf{H}_{pp}$ is well conditioned. (b) In smoke-filled, visually degraded scenes, the visual detector identifies a decrease in transmittance; the system introduces radar Doppler-velocity constraints $\mathbf{z}_{R_V}$ and switches to the LiDAR+IMU+Visual+Radar (LIVR) mode. (c) In feature-sparse, geometrically degenerate corridors, the geometric detector identifies an under-constrained direction in $x$-axis translation (the ellipsoid defined by the eigenvectors of $\mathbf{H}_{pp}$ becomes degenerate); the system introduces wheel-odometry velocity constraints $\mathbf{z}_{W}$ and switches to the LiDAR+IMU+Visual+Wheel (LIVW) mode.}
    \label{Fig3-Dengerate_detection}
    \vspace{0cm}
\end{figure}

This section presents an observability-analysis-driven strategy for degradation detection and adaptive fusion-mode switching, as illustrated in Fig.~\ref{Fig3-Dengerate_detection}. The module comprises two components: degradation detection and adaptive fusion mode switching. It introduces additional sensor constraints and switches fusion modes online to improve SLAM robustness in challenging environments.

We consider two forms of degradation: (i) visual degradation caused by smoke and dust, which substantially reduces image contrast and obscures texture information; and (ii) geometric degradation caused by long, straight corridors, in which point-cloud registration becomes under-constrained along certain degrees of freedom. Accordingly, FIRE-LIVWO employs two complementary detectors. \textbf{(1) Visual degradation detector (camera-centric):} determines whether the current image is in a smoke-induced low-visibility state and decides whether to suppress visual photometric constraints while strengthening mmWave radar constraints. \textbf{(2) Geometric degradation detector (LiDAR-centric):} analyzes the Hessian matrix associated with the point-to-plane residuals $\mathbf{r}_{L}$ and $\mathbf{r}_{R_G}$ to identify under-constrained directions in geometric registration and decides whether to introduce or strengthen wheel-odometry constraints. Finally, the system selects the fusion mode based on the outputs of both detectors and integrates it into the IESKF iterated update in (\ref{eq_4_constrains}).

\subsubsection{Visual Degradation Detector}\label{chap5-3-1-visual-detector}

Smoke- and dust-induced image degradation can be approximated by the classical atmospheric scattering model~\cite{narasimhan2002vision}:
\begin{equation}
    I(\mathbf u) = J(\mathbf u)\,t(\mathbf u) + A\,(1-t(\mathbf u)),
    \label{eq_13_atmosphere_model}
\end{equation}
where $I(\mathbf u)$ is the observed pixel intensity, $J(\mathbf u)$ is the haze-free scene radiance, $A$ is the airlight, and $t(\mathbf u)=\exp(-\beta d(\mathbf u))\in(0,1]$ is the transmittance. Here, $\beta$ is the scattering coefficient and $d(\mathbf u)$ is the scene depth. Denser smoke yields smaller transmittance, lower contrast, and fewer visible details, thereby reducing the amount of usable visual information.

For real-time smoke detection, we estimate the transmittance $\hat t(\mathbf u)$ using the dark channel prior~\cite{he2010single} and define the visual observability score $\mathcal{O}_V$ for the current frame as the average transmittance:
\begin{equation}
    \mathcal O_V \triangleq \bar t_k = \frac{1}{|\mathcal U|}\sum_{\mathbf u\in\mathcal U}\hat t(\mathbf u) \in (0,1] \quad,
    \label{eq_15_transmittance_avg}
\end{equation}
where $\mathcal U$ denotes the set of sampled pixels. A larger $\mathcal{O}_V$ indicates higher transmittance and richer image details, and therefore more reliable visual photometric constraints in (\ref{eq_11_visual_residual}). We define the visual degradation criterion as
\begin{equation}
    \mathcal D_V =
    \begin{cases}
    1, & \mathcal O_V \le \mathcal O_V^{th},\\
    0, & \text{otherwise},
    \end{cases}
    \label{eq_17_visual_obs}
\end{equation}
where $\mathcal O_V^{th}$ is the threshold for visual degradation. When $\mathcal D_V=1$, the system is in a visually degraded state. In this case, the sparse direct photometric residual in (\ref{eq_11_visual_residual}) becomes less reliable due to contrast loss and scattering-induced attenuation. The system should therefore down-weight the visual term or disable visual updates and instead rely on Doppler-velocity constraints to maintain state estimation.

\subsubsection{Geometric Degradation Detector}\label{chap5-3-2-geometric-detector}

Geometric degradation essentially arises when the point-cloud registration problem becomes under-constrained along certain degrees of freedom, which is equivalent to the existence of an null space in the corresponding Hessian matrix. By analyzing the principal directions of the Hessian in point-to-plane ICP optimization, the under-constrained directions can be identified, and unreliable updates along those directions can be suppressed accordingly.

Let $\mathbf J_G$ denote the stacked Jacobians of the LiDAR and radar point-to-plane residuals. We define the geometric Hessian under the weighted least-squares formulation as $\mathbf H_G \triangleq \mathbf J_G^{\top}\mathbf W_G \mathbf J_G$, where
$\mathbf{W}_G=\mathrm{blkdiag}\Big((\mathbf P_L^1)^{-1},\ldots,(\mathbf P_L^{N_L})^{-1},(\mathbf P_{R_G}^1)^{-1},\ldots,(\mathbf P_{R_G}^{N_{R_G}})^{-1}\Big)$.

From $\mathbf H_G\in\mathbb R^{18\times18}$, we extract the sub-block corresponding to
$ \delta\mathbf x=[\delta{}^G\boldsymbol\theta_I^{\top},\ \delta{}^G\mathbf p_I^{\top}]^{\top}\in\mathbb R^{6}$: $\mathbf H_{pose}= \begin{bmatrix} \mathbf H_{rr} & \mathbf H_{rp}\\ \mathbf H_{pr} & \mathbf H_{pp} \end{bmatrix}\in\mathbb R^{6\times6}.$
We then perform eigen-decomposition on the rotational sub-matrix $\mathbf H_{rr}$ and the translational sub-matrix $\mathbf H_{pp}$, respectively:
\begin{equation}
    \mathbf H_{pp}=\mathbf V_p \boldsymbol{\Lambda}_p \mathbf V_p^{\top},\quad \mathbf H_{rr}=\mathbf V_r \boldsymbol{\Lambda}_r \mathbf V_r^{\top},
    \label{eq_21_geometric_eigen_matrix}
\end{equation}
where $\boldsymbol {\Lambda}_p=\mathrm{diag}(\lambda_{p_1},\lambda_{p_2},\lambda_{p_3})$ with $\lambda_{p_1}\ge\lambda_{p_2}\ge\lambda_{p_3}$, and the rotational part is defined analogously. We characterize local observability using the relative strength of eigenvalues:
\begin{equation}
    s_p \triangleq \left\vert \frac{\lambda_{p_1}}{\lambda_{p_3}+\epsilon}\right\vert,\qquad
    s_r \triangleq \left\vert \frac{\lambda_{r_1}}{\lambda_{r_3}+\epsilon}\right\vert,
    \label{eq_21_geometric_eigen_score}
\end{equation}
and define the geometric observability score as $\mathcal O_G = \max(s_p,s_r)$. The geometric degradation criterion is then given by
\begin{equation}
    \mathcal D_G =
    \begin{cases}
    1, & \mathcal O_G \ge \max(s_p^{th},s_r^{th}),\\
    0, & \text{otherwise},
    \end{cases}
    \label{eq_23_geo_degradation_condition}
\end{equation}
where $s_p^{th}$ and $s_r^{th}$ are the degradation thresholds for translation and rotation, respectively. The eigenvectors $\mathbf v_{p_3}$ and $\mathbf v_{r_3}$ associated with the smallest eigenvalues indicate the degenerate directions. This interpretation helps explain common structural degeneracies, such as under-constrained translation along the corridor direction or under-constrained rotation about a specific axis.

\subsubsection{Adaptive Fusion-Mode Switching}\label{chap5-3-3-degradation-detection-switching}

Based on the detection outputs, we develop an adaptive fusion-mode switching mechanism. The visual and geometric detectors first identify the degradation type and compute the observability scores $\mathcal{O}_V$ and $\mathcal{O}_G$ to quantify degradation severity. When a score crosses its corresponding threshold, the system activates additional constraints—mmWave radar Doppler-velocity constraints or wheel-odometry velocity constraints—to enhance robustness.

As illustrated in Fig.~\ref{Fig3-Dengerate_detection}, the switching strategy consists of three stages.
\textbf{(a) Stage I: LIV mode (nominal scenarios).}
As shown in Fig.~\ref{Fig3-Dengerate_detection}-(a), when $\mathcal{D}_V=0$ and $\mathcal{D}_G=0$, the system operates in feature-rich corridor scenes, where the ellipsoid induced by $\mathbf{H}_{pp}$ is well conditioned. The estimator fuses LiDAR measurements $\mathbf{z}_L$, visual measurements $\mathbf{z}_C$, and IMU measurements to localize the platform.
\textbf{(b) Stage II: LIVR mode (visual degradation).}
As shown in Fig.~\ref{Fig3-Dengerate_detection}-(b), when $\mathcal{D}_V=1$ and $\mathcal{D}_G=0$, the visual detector indicates smoke-induced visual degradation. The system activates radar Doppler-velocity constraints $\mathbf{z}_{R_V}$ and switches to the LIVR mode.
\textbf{(c) Stage III: LIVW mode (geometric degradation).}
As shown in Fig.~\ref{Fig3-Dengerate_detection}-(c), when $\mathcal{D}_V=0$ and $\mathcal{D}_G=1$ (i.e., $\mathcal{O}_G \ge \mathcal{O}_G^{th}$), the geometric detector indicates that registration is under-constrained along at least one direction (e.g., translation along the $x$-axis). The system activates wheel-odometry velocity constraints $\mathbf{z}_{W}$ and switches to the LIVW mode to compensate for missing constraints along degenerate directions.

When the system simultaneously encounters visual and geometric degradation (e.g., a long, straight corridor filled with dense smoke), it activates both radar Doppler-velocity constraints and wheel-odometry velocity constraints, thereby entering the LIVRW mode (LiDAR+IMU+Visual+Radar+Wheel) to maintain robust state estimation under dual degradation.

\begin{figure}[t]
    \centering
    \includegraphics[width=1\columnwidth]{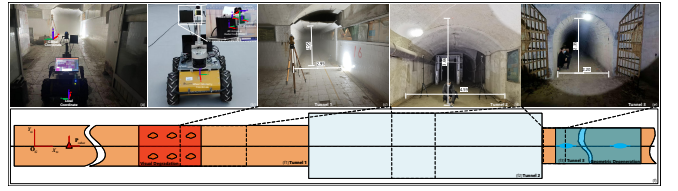}
    \caption{Experimental environment in underground coal-mine tunnels. (a) Deployment of the Husky robot and the total station; (b) sensor configuration on the Husky platform, including LiDAR, camera, 4D mmWave radar, IMU, and wheel odometer; (c)--(e) environments, dimensions, and texture details of Tunnel1, Tunnel2, and Tunnel3, respectively; (f) locations exhibiting the most severe visual and geometric degradations (f1, f3).}
    \label{Fig4-field-exp-setup}
\end{figure}

\section{Experiment}\label{chap6-experiment}

\subsection{Experiment Setup}\label{chap6-1-experiment}
\begin{figure*}
    \centering
    \includegraphics[width=1\textwidth]{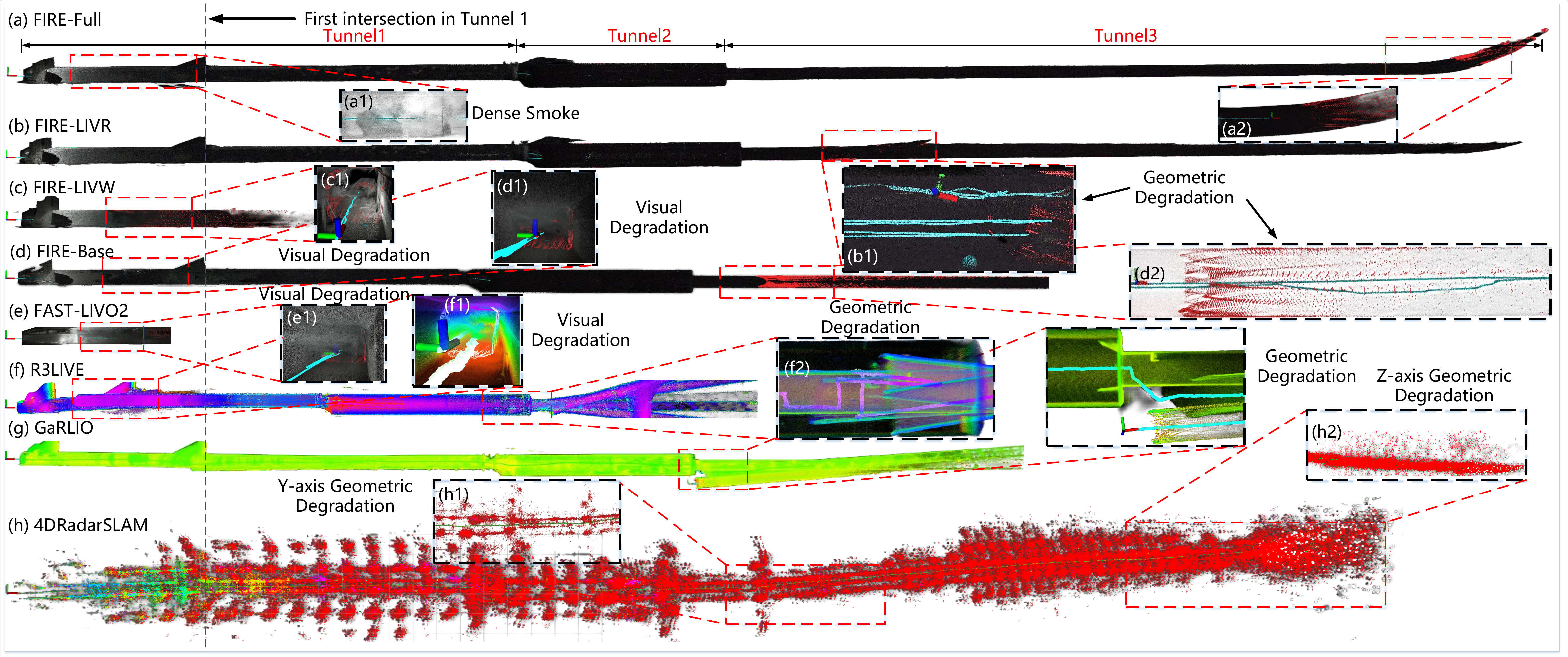}
    \caption{Comparison of mapping results of different algorithms in underground coal-mine tunnels. (a) FIRE-Full, (b) FIRE-LIVR, (c) FIRE-LIVW, (d) FIRE-Base, (e) FAST-LIVO2, (f) R3LIVE, (g) GaRLIO, and (h) 4DRadarSLAM. In Tunnel1, FIRE-LIVW and FAST-LIVO2 fail due to smoke-induced visual degradation. In Tunnel3, FIRE-LIVR, FIRE-Base, R3LIVE, and GaRLIO suffer from severe geometric degeneration. 4DRadarSLAM exhibits significant drift along the $y$- and $z$-axes. Only FIRE-Full successfully reaches the end of the tunnel.}
    \label{Fig5-Mapping_results}
\end{figure*}

\begin{figure}[htpb]
    \centering
    \includegraphics[width=1\columnwidth]{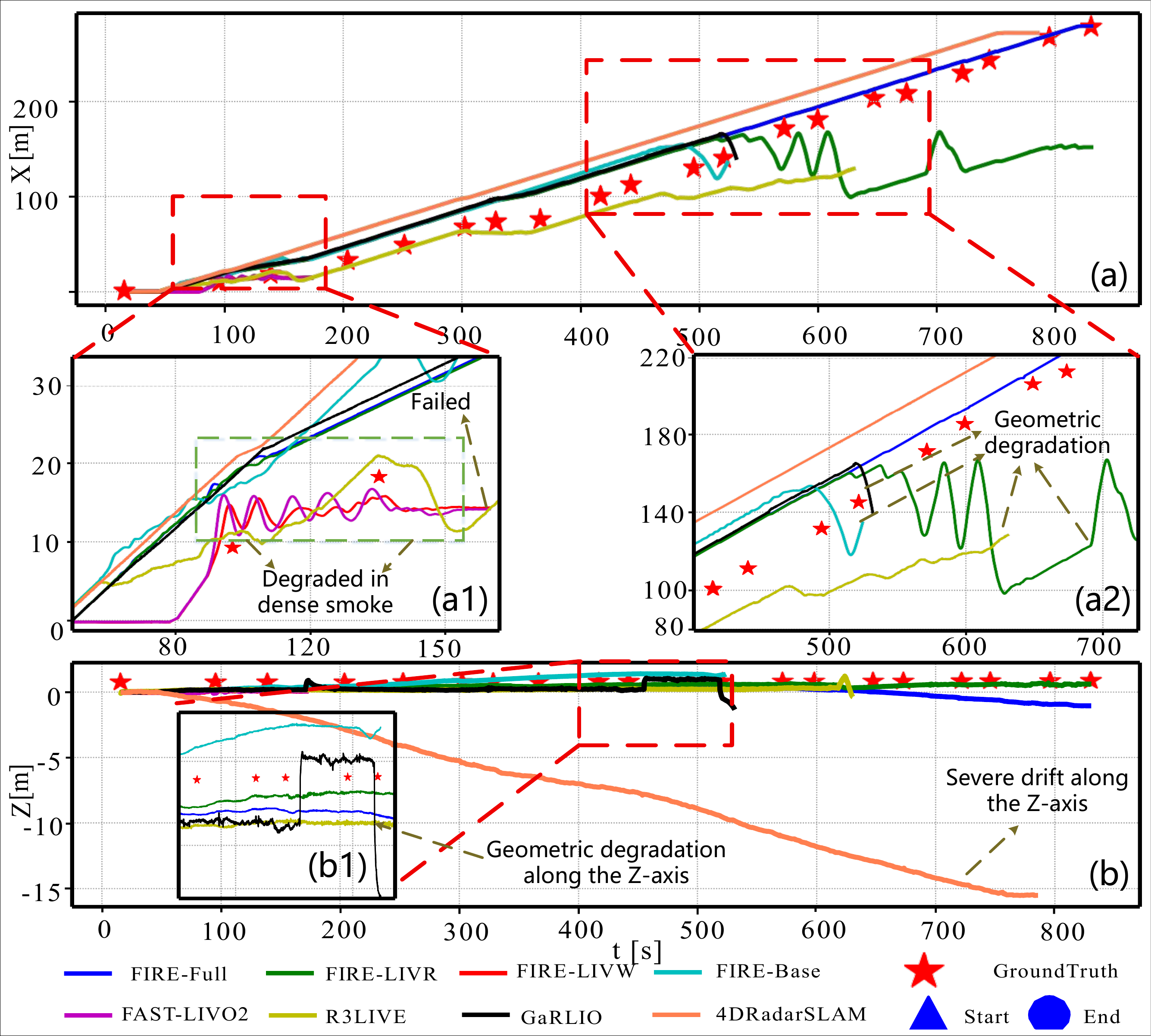}
    \caption{Time evolution of the $x$- and $z$-axis trajectory components for different methods. (a) and (b) show the $x$- and $z$-axis components of FIRE-Full, FIRE-LIVR, FIRE-LIVW, FIRE-Base, FAST-LIVO2, R3LIVE, GaRLIO, and 4DRadarSLAM, respectively. (a1) and (a2) provide zoomed-in views of the $x$-axis components, and (b1) shows a zoomed-in view of the $z$-axis components.}
    \label{Fig6-Trajectory_results}
    \vspace{0cm}
\end{figure}


\textbf{Hardware and software platform:} We conducted field experiments in severely degenerate underground coal-mine tunnels. The mobile platform is a Husky A200 equipped with a Livox AVIA LiDAR (with a built-in IMU), a Hikvision camera (MVS-CU013-A0UC), an Oculii Eagle 4D mmWave radar, and the chassis wheel odometer. The LiDAR--camera and radar--IMU extrinsics were carefully calibrated, and a hardware synchronizer time-synchronized the LiDAR, IMU, camera, and mmWave radar. Ground-truth points were measured using a total station. All computations were performed on an industrial PC with an Intel i7 CPU, 32~GB DDR4 RAM, an NVIDIA GeForce GTX 1050Ti GPU, and a 512~GB solid-state drive (SSD).

\textbf{Experimental design:} To comprehensively evaluate the proposed method, we designed experiments to assess trajectory accuracy as well as degradation detection and adaptive fusion-mode switching. We compare four variants of FIRE-LIVWO—FIRE-Base (LiDAR+IMU+Visual), FIRE-LIVR (LiDAR+IMU+Visual+Radar), FIRE-LIVW (LiDAR+IMU+Visual+Wheel), and FIRE-Full (LIVRW)—against FAST-LIVO2 (LIV), R3LIVE (LIV), GaRLIO (RLI), and 4DRadarSLAM (Radar).

\subsection{Real-world Experiment}\label{chap6-2-experiment}

This section evaluates the proposed method in real underground coal-mine tunnels in terms of trajectory accuracy, degradation detection and adaptive fusion-mode switching. Fig.~\ref{Fig4-field-exp-setup}-(a,f) shows the deployment of the robot and the total station in the tunnel. We define the total-station center $\mathbf{O}_G$ as the origin of the world frame, and set the robot initial position to $\mathbf{P}_{robot}=(-6.261, 1.222 ,0.268)$. Fig.~\ref{Fig4-field-exp-setup}-(b) illustrates the sensor layout on the robot platform. The robot moves forward at a constant speed of 0.3~m/s, with a maximum angular speed of 0.2~rad/s. Fig.~\ref{Fig4-field-exp-setup}-(c)--(e) depict the environments of Tunnel1, Tunnel2, and Tunnel3, respectively: Tunnel1 contains relatively rich features but suffers from severe smoke and dust; Tunnel2 is smoke-free but feature-sparse; and Tunnel3 is extremely feature-sparse and highly repetitive. These conditions pose substantial challenges for SLAM: (1) smoke occlusion significantly degrades visual performance, making Tunnel1 prone to visual degradation; (2) long, straight structures weaken LiDAR geometric constraints, particularly during motion aligned with the tunnel direction; and (3) repetitive geometry reduces distinctiveness, making Tunnel3 prone to geometric degradation. Despite these challenges, our system achieves favorable localization accuracy and robustness.

\subsubsection{Trajectory Accuracy Analysis}\label{chap6-2-1-traj-experiment}

Figures~\ref{Fig5-Mapping_results} and~\ref{Fig6-Trajectory_results} present the mapping and trajectory results for all methods, respectively. Because GPS is unavailable in underground coal-mine tunnels, we evaluate trajectory accuracy using 20 ground-truth points, $\mathbf{p}_{gt}^i$, measured by a total station and the corresponding estimated points, $\mathbf{p}_{es}^i$, produced by each method. The average localization error is defined as $\mathrm{AvgErr}=\sum_{i=0}^{N-1} \frac{||\mathbf{p}_{es}^i-\mathbf{p}_{gt}^i||}{N}$.(\emph{Remark:} if a method crashes or numerically diverges during the sequence and cannot output the full trajectory, it is marked as \emph{failure} and denoted by $\times$.)

The baseline methods exhibit varying degrees of anomalies and failures under these extreme conditions. In Tunnel1, FAST-LIVO2 and FIRE-LIVW, which rely on visual and LiDAR features, fail shortly after entering the tunnel due to smoke-induced visual degradation (Fig.~\ref{Fig5-Mapping_results}-(e,c) and Fig.~\ref{Fig6-Trajectory_results}-(a,a1)) and therefore cannot output complete trajectories. FIRE-Base and R3LIVE operate over parts of the route but show visual degradation in Tunnel1 (Fig.~\ref{Fig5-Mapping_results}-(d1,f1) and Fig.~\ref{Fig6-Trajectory_results}-(a1)). They also exhibit trajectory deviation and map misalignment in Tunnel3 (Fig.~\ref{Fig5-Mapping_results}-(d2,f2) and Fig.~\ref{Fig6-Trajectory_results}-(a2)). Methods equipped with 4D mmWave radar are more resilient to visual degradation. GaRLIO and FIRE-LIVR mitigate smoke-induced visual degradation in Tunnel1, but still suffer pronounced geometric degeneration in the highly repetitive environment of Tunnel3 (Fig.~\ref{Fig6-Trajectory_results}-(a2) and Fig.~\ref{Fig5-Mapping_results}-(g1,b1)). Although 4DRadarSLAM does not crash, owing to the strong penetration capability of mmWave radar (Fig.~\ref{Fig5-Mapping_results}-(h)), its limited ranging accuracy and sparse point clouds lead to severe drift of up to 15~m along the $y$- and $z$-axes (Fig.~\ref{Fig6-Trajectory_results}-(b)). In contrast, the full system, FIRE-Full, achieves the best robustness and accuracy: it overcomes smoke-induced visual failures in Tunnel1 by leveraging radar and suppresses geometric degeneration in Tunnel3 by adaptively fusing wheel-odometry constraints, ultimately reaching the end of the tunnel (Fig.~\ref{Fig5-Mapping_results}-(a) and Fig.~\ref{Fig6-Trajectory_results}-(a)).

To quantitatively assess localization performance, Table~\ref{tbl2:localization_error} reports $\mathrm{AvgErr}$ for all methods. FIRE-Full achieves an $\mathrm{AvgErr}$ of 5.677~m, substantially outperforming all baselines. FIRE-LIVR and FIRE-Base yield errors of 15.420~m and 23.962~m, respectively, indicating that removing either radar or wheel constraints significantly degrades accuracy. R3LIVE (31.596~m), GaRLIO (17.212~m), and 4DRadarSLAM (45.453~m) exhibit large drift under different degradation mechanisms, whereas FIRE-LIVW and FAST-LIVO2 fail due to visual degradation. Overall, these results highlight the importance of multi-modal fusion: omitting any modality leads to a pronounced performance drop.

\begin{table}[htpb]
    \centering
    \caption{Average localization error analysis (m). }\label{tbl2:localization_error}
    \begin{threeparttable}
        \begin{tabularx}{\linewidth}{@{}l *{8}{>{\centering\arraybackslash}X}@{}}
            \toprule
             & \rotatebox{30}{\makecell{FIRE-\\Full}} 
             & \rotatebox{30}{\makecell{FIRE-\\LIVR}} 
             & \rotatebox{30}{\makecell{FIRE-\\LIVW}} 
             & \rotatebox{30}{\makecell{FIRE-\\Base}} 
             & \rotatebox{30}{\makecell{FAST-\\LIVO2}}
             & \rotatebox{30}{\makecell{R3LIVE}} 
             & \rotatebox{30}{\makecell{GaRLIO}} 
             & \rotatebox{30}{\makecell{4DRadar-\\SLAM}} \\
            \midrule
		 $\mathrm{AvgErr}$ &  \textbf{5.677}	 & \underline{15.420} & $\times$ & 23.962 & $\times$ & 31.596 & 17.212 & 45.453   \\            
         \bottomrule
        \end{tabularx}
        \begin{tablenotes}
            \item \textbf{Bold} indicates the best accuracy; \underline{underline} indicates the second best. $\times$ indicates failure.
        \end{tablenotes}
    \end{threeparttable}
\end{table}

\subsubsection{Analysis of Degradation Detection and Adaptive Fusion-Mode Switching}\label{chap6-2-2-degradation-experiment}

\begin{figure}[t]
    \centering
    \includegraphics[width=1\columnwidth]{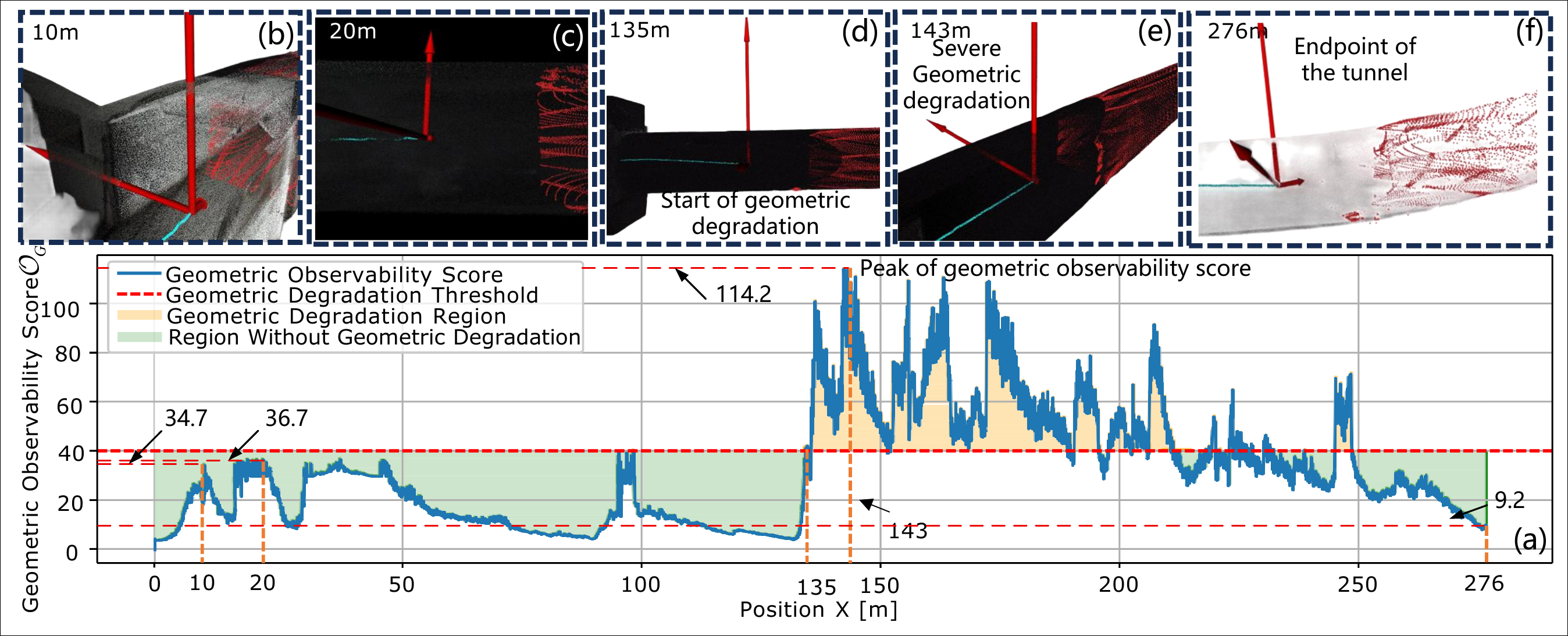}
    \caption{Geometric observability score $\mathcal O_G$ and visualization of the eigenvectors of $\mathbf H_{pp}$ (red arrows indicate eigen-directions, and their lengths represent eigenvalue magnitudes). (a) $\mathcal O_G$ versus the robot $x$-position; (b)--(f) eigenvectors at key locations: inside Tunnel1 ($x=10,20$~m, $\mathcal O_G=34.7, 36.7$), the onset of geometric degradation ($x=135$~m, $\mathcal O_G=40$), the most severe geometric degradation ($x=143$~m, $\mathcal O_G=114.2$), and the end of Tunnel3 ($x=276$~m, $\mathcal O_G=9.2$). The red dashed line denotes the threshold $\mathcal{O}_G^{th}=40$.}
    \label{Fig7-Geo_observability}
\end{figure}

\begin{figure}[h]
    \centering
    \includegraphics[width=1\columnwidth]{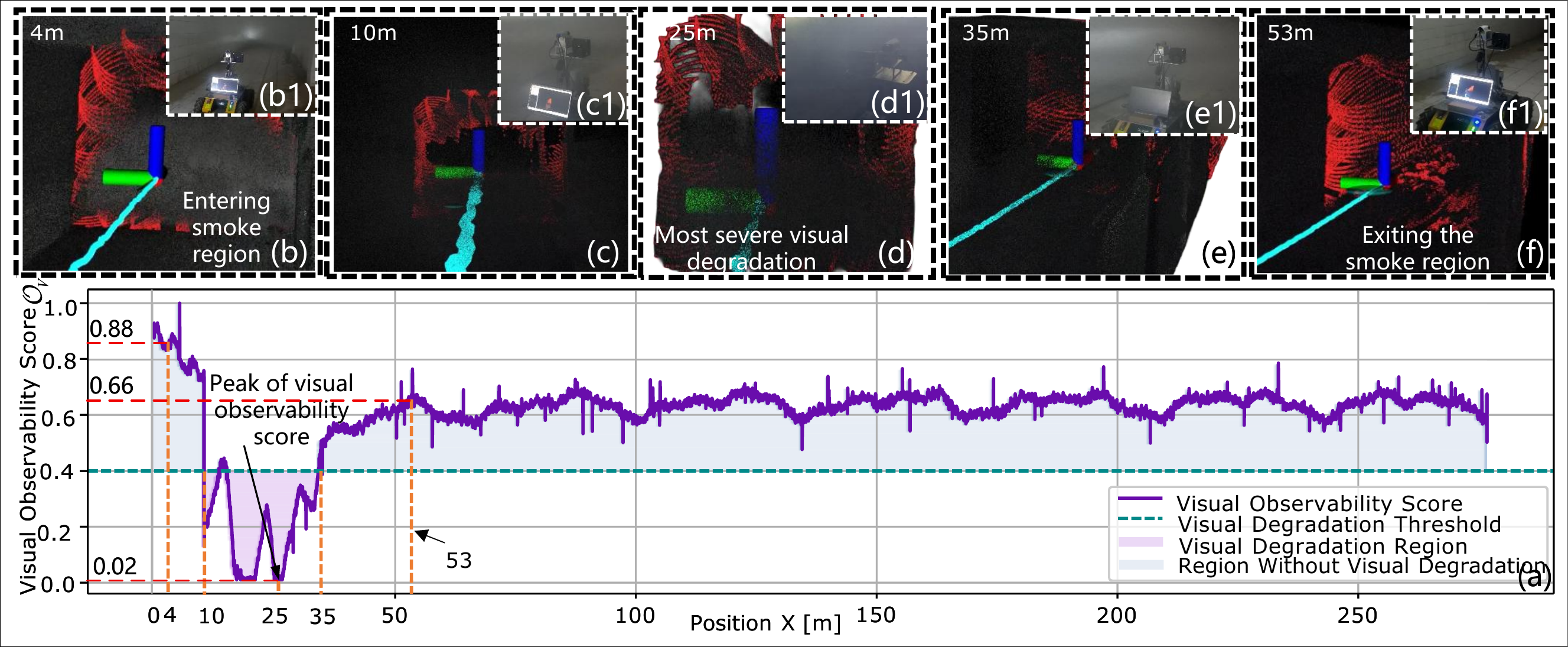}
    \caption{Visual observability score $\mathcal{O}_V$ and system behavior in smoke scenarios. (a) $\mathcal{O}_V$ versus the robot $x$-position; (b,b1)--(f,f1) corresponding odometry behavior and in-situ photos: initial clear region ($x=4$~m, $\mathcal{O}_V=0.88$), entering the smoke region ($x=10$~m, $\mathcal{O}_V=0.66$), densest smoke region ($x=25$~m, $\mathcal{O}_V=0.02$), weakening smoke region ($x=35$~m, $\mathcal{O}_V=0.40$), and smoke-free region ($x=53$~m, $\mathcal{O}_V=0.66$). The red dashed line denotes the threshold $\mathcal{O}_V^{th}=0.4$.}
    \label{Fig8-Visual_observability}
\end{figure}

This section further validates the effectiveness of the proposed visual and geometric degradation detectors. Figures~\ref{Fig7-Geo_observability} and~\ref{Fig8-Visual_observability} show how the geometric observability score $\mathcal{O}_G$ and the visual observability score $\mathcal{O}_V$ (for FIRE-Full) vary with the robot $x$-position in geometrically constrained (Tunnel3) and visually constrained (Tunnel1) scenarios, respectively, together with visualizations at representative locations.

\textbf{Geometric degradation detection:}
As shown in Fig.~\ref{Fig7-Geo_observability}, when the robot traverses the feature-rich Tunnel1 ($x\in[0,135]$~m), $\mathcal{O}_G$ remains below the threshold $\mathcal{O}_G^{th}$. After entering Tunnel3, the environment transitions abruptly to a textureless, long, straight corridor; consequently, $\mathcal{O}_G$ increases rapidly and exceeds $\mathcal{O}_G^{th}$, reaching a peak of 114.2 at $x=143$~m. At this location, the smallest eigenvalue of $\mathbf{H}_{pp}$ becomes extremely small (Fig.~\ref{Fig7-Geo_observability}-(d,e)), indicating that point-cloud registration is severely under-constrained along the corresponding direction. This under-constraint causes pronounced drift in the $x$-axis direction for R3LIVE, GaRLIO, and FIRE-LIVR in this region (Fig.~\ref{Fig6-Trajectory_results}-(a2)). In contrast, FIRE-Full monitors $\mathcal{O}_G$ online and, once $\mathcal{D}_G=1$ is detected, activates tightly coupled wheel-odometry constraints to prevent divergence along the degenerate direction, thereby maintaining stable estimates throughout the geometrically degenerate interval (highlighted in yellow in Fig.~\ref{Fig7-Geo_observability}-(a)).

\textbf{Visual degradation detection:}
As shown in Fig.~\ref{Fig8-Visual_observability}, in the initial clear region of Tunnel1 ($x=4$~m), $\mathcal{O}_V=0.88$ and the system localizes using tightly-coupled visual--LiDAR--inertial constraints. After the robot enters the smoke region, scattering causes a sharp drop in transmittance; $\mathcal{O}_V$ decreases rapidly and falls below the threshold $\mathcal{O}_V^{th}$ at $x=10$~m, triggering visual degradation ($\mathcal{D}_V=1$). In the densest smoke region ($x=25$~m), $\mathcal{O}_V$ drops to 0.02, and camera observations become nearly invalid (Fig.~\ref{Fig8-Visual_observability}-(d,d1)). FIRE-LIVW and FAST-LIVO2, lacking online observation-quality assessment, introduce erroneous visual residuals in this region, leading to divergence and failure (Fig.~\ref{Fig6-Trajectory_results}-(a1)), whereas FIRE-Base and R3LIVE do not crash but exhibit pronounced visual degradation. By contrast, once FIRE-Full detects $\mathcal{D}_V=1$, it adaptively down-weights the visual term and switches to radar-dominant observations, leveraging radar penetration to compensate for missing exteroceptive information and stably traverse the smoke region. When $\mathcal{O}_V$ recovers to 0.66 at $x=53$~m, the system reactivates vision for state estimation (Fig.~\ref{Fig8-Visual_observability}-(f,f1)).


Overall, these results show that $\mathcal{O}_V$ and $\mathcal{O}_G$ reliably indicate visual failures and geometric under-constraints in real underground coal-mine tunnels. FIRE-Full performs two key online switches (visual-degradation interval: LIV$\rightarrow$LIVR$\rightarrow$LIV; geometric-degradation interval: LIV$\rightarrow$LIVW$\rightarrow$LIV), achieving the best accuracy and robustness under extreme mixed-degradation conditions.


\section{Conclusion}\label{chap7-conclusion}

This paper presents FIRE-LIVWO, a tightly coupled multi-modal odometry framework designed for extreme underground coal-mine scenarios characterized by large-scale environments, severe smoke and dust interference, and geometric degeneration in long tunnels. The proposed method jointly models shared map elements across LiDAR, radar, and vision within a unified VoxelMap and, within a single filter update, fuses LiDAR--radar point-to-plane geometric residuals with sparse direct visual photometric residuals in a unified formulation. We further incorporate pointwise Doppler-velocity constraints from 4D mmWave radar to improve observability and resilience to visual failure in smoke-filled environments. In addition, wheel odometry is tightly coupled into the estimator via non-holonomic constraints (NHC) and online lever-arm compensation, thereby reducing drift induced by long, straight corridor degeneracy. We also develop an observability-analysis-based degradation detection and adaptive fusion strategy that identifies smoke interference and geometric underconstraints online and dynamically switches fusion modes to maintain continuous and stable state estimation across scenarios. Comparative and ablation studies in real underground coal mines demonstrate that FIRE-LIVWO can robustly traverse severely degraded environments, including dense smoke and long, featureless tunnels, achieving superior localization accuracy and robustness compared with existing methods, with an average localization error  $\mathrm{AvgErr}$ of 5.677~m.

\bibliographystyle{IEEEtranBST/IEEEtran}
\bibliography{IEEEtranBST/ref_fire_livwo}

\end{document}